%% file: main.tex
\documentclass[letterpaper, 10 pt, journal, twoside]{IEEEtran}
\IEEEoverridecommandlockouts                              

\usepackage{amsmath,bm,amsfonts,cases}
\usepackage{shortcuts,subfiles,balance}
\usepackage{booktabs,xcolor}
\usepackage{graphicx}
\usepackage{float}
\usepackage{amsthm}
\usepackage{amssymb}
\usepackage{multirow}
\usepackage{url}
{                                                       
    \theoremstyle{plain}
    
    \newtheorem{proposition}{Proposition}

}

\begin{document}
\title{CLF-RL: Control Lyapunov Function Guided Reinforcement Learning}

\author{%
Kejun Li$^{*,1}$, Zachary Olkin$^{*,2}$, Yisong Yue$^{3}$, Aaron D. Ames$^{2}$%
\thanks{Manuscript received: Aug. 12, 2025; Revised Nov. 13, 2025; Accepted Dec. 17, 2025.}%
\thanks{This paper was recommended for publication by Editor Jens Kober upon evaluation of the Associate Editor and Reviewers’ comments.
This research is supported by Technology Innovation Institute (TII) and the National Science Foundation Graduate Research Fellowship. Supplementary video available: \protect\url{https://youtu.be/KxSwKXjjtko}}%
\thanks{* denotes equal contribution. $^{1}$Kejun Li is with the Department of Biology and Biological Engineering, California Institute of Technology, Pasadena, CA, USA 
{\tt kli5@caltech.edu}}%
\thanks{$^{2}$Zachary Olkin and Aaron D. Ames are with the Department of Control and Dynamical Systems, California Institute of Technology, CA, USA 
{\tt \{zolkin, ames\}@caltech.edu}}%
\thanks{$^{3}$Yisong Yue is with the Department of Computing and Mathematical Sciences, California Institute of Technology, Pasadena, CA, USA 
{\tt yyue@caltech.edu}}%
\thanks{Digital Object Identifier (DOI): see top of this page}%
}


\maketitle

\begin{abstract}
Reinforcement learning (RL) has shown promise in generating robust locomotion policies for bipedal robots, but often suffers from tedious reward design and sensitivity to poorly shaped objectives. In this work, we propose a structured reward shaping framework that leverages model-based trajectory generation and control Lyapunov functions (CLFs) to guide policy learning. We explore two model-based planners for generating reference trajectories: a reduced-order linear inverted pendulum (LIP) model for velocity-conditioned motion planning, and a precomputed gait library based on hybrid zero dynamics (HZD) using full-order dynamics. These planners define desired end-effector and joint trajectories, which are used to construct CLF-based rewards that penalize tracking error and encourage rapid convergence. This formulation provides meaningful intermediate rewards, and is straightforward to implement once a reference is available. Both the reference trajectories and CLF shaping are used only during training, resulting in a lightweight policy at deployment. We validate our method both in simulation and through extensive real-world experiments on a Unitree G1 robot. CLF-RL demonstrates significantly improved robustness relative to the baseline RL policy and better performance than a classic tracking reward RL formulation.
\end{abstract}



\subfile{sections/introduction}
\subfile{sections/background}

\subfile{sections/reward_design}
\subfile{sections/framework}

\subfile{sections/results}

\section{Conclusion}
We presented CLF-RL, a structured reward shaping framework that integrates reference trajectory tracking with control Lyapunov functions (CLFs). By embedding CLF-based objectives directly into the reinforcement learning reward, our approach replaces heuristic reward design with a theoretically grounded stability metric. We showed that incorporating the CLF decay condition improves tracking performance over baseline tracking-only rewards. CLF-based policies exhibit superior robustness across a range of perturbations in both simulation and hardware, demonstrating increased reliability.

Our framework is modular and extensible. We demonstrated its compatibility with both reduced-order (H-LIP) and full-order (HZD) reference generators, though other trajectory planners could be incorporated. While our training only used steady-state reference motions, the learned policies generalize well to transient motions—highlighting the framework’s flexibility. We validated the real-world robustness of the approach through extended outdoor trials, showing consistent and stable walking performance across varied flat-ground terrains. Finally, while our rewards are derived from CLFs that are theoretically sound for the full-order system, they are applied as guidance rather than hard constraints, leaving formal stability guarantees as a direction for future work.

\bibliographystyle{IEEEtran}
\balance
\bibliography{IEEEabrv, References}

\end{document}

%% file: sections/introduction.tex
\section{Introduction}

\IEEEPARstart{A}{chieving} stable and agile bipedal locomotion remains a central challenge in robotics due to the high-dimensional, underactuated, and hybrid nature of legged dynamics, which combine continuous motion and discrete events such as foot-ground impacts. This hybrid behavior poses significant challenges for designing controllers that combine precise coordination with robustness for real-world deployment.

Traditional model-based control frameworks address these challenges through principled formulations such as trajectory optimization. Methods based on hybrid zero dynamics (HZD) \cite{westervelt2003hybrid} and feedback-linearization have enabled offline generation of stable walking gaits under physical and task constraints \cite{reher2020algorithmic}. These techniques provide strong guarantees on stability and constraint satisfaction but require solving large-scale nonlinear programs that are computationally demanding and sensitive to model mismatch, limiting their practicality for deployment. To enable faster and more reactive planning, many approaches leverage reduced-order models such as the linear inverted pendulum (LIP) and its variants \cite{kajita2003biped,kajita_3d_2001}, which simplify the dynamics by focusing on the center of mass and foot placement. While computationally efficient and suitable for stabilization and mid-level planning, these models rely on strong assumptions such as massless legs and constant center of mass height, which limit their ability to generate highly dynamic and agile whole body motion.

\begin{figure}[t!]
    \centering
    \includegraphics[width=1.0\linewidth]{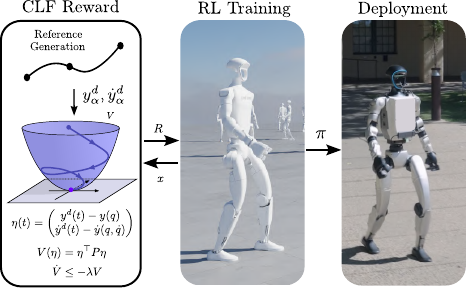}
    \caption{Overview of our approach. A reference generator produces target trajectories, which are used to construct a CLF-based reward. An RL policy is trained in simulation with this reward and deployed on a real humanoid robot.}
    \label{fig:hero_fig}
    \vspace{-6mm}
\end{figure}

Recent advances in computation and GPU-parallelizable physics simulators have made reinforcement learning (RL) a promising alternative for locomotion control. By offloading the heavy computation to offline training in massively parallel simulations, RL enables lightweight policy inference during online deployment to hardware \cite{lee2020learning,xie2020learning,rudin_learning_2022,zhuang_humanoid_2024,li_reinforcement_2024}. However, applying RL to bipedal locomotion remains challenging due to its reliance on heuristic reward design, which is tedious to construct, sensitive to tuning, and, if poorly shaped, can lead to unstable gaits, prolonged training, and poor sim-to-real transfer.
A strategy to alleviate the burden of this process is to embed model-based reference trajectories directly into the RL reward. This idea has been explored using reduced-order models \cite{batke_optimizing_2022,green_learning_2021}, full-order models \cite{liu_opt2skill_2025}, and with the HZD framework \cite{li_reinforcement_2021}. \cite{lee_integrating_2024} proposes a different way to incorporate a LIP controller into RL by generating desired footstep locations during training and using it to provide feedback control for rough terrain.

While prior model-guided approaches improve structure during training, they typically reward only the instantaneous tracking error, treating two states with the same error magnitude equally—even if one is actively converging toward the reference and the other is diverging. This can cause policies to undervalue transient corrections. To address these limitations, we propose a reward shaping framework built on control Lyapunov functions (CLFs), a fundamental tool in nonlinear control theory for generating certifiably stable controllers \cite{sontag_lyapunov-like_1983,sontag_universal_nodate,artstein_stabilization_1983}. CLFs have previously been used in bipedal locomotion, often in conjunction with HZD, to formally guarantee the stability of periodic gaits \cite{ames_rapidly_2014} and have also been integrated into learning-based methods in other domains \cite{westenbroek_combining_2021,choi_reinforcement_2020}. For instance, \cite{westenbroek_lyapunov_2022} introduces a reward-reshaping method that incorporates a candidate CLF for fine-tuning policies using minimal hardware data. 

Our approach embeds both the CLF and its associated decrease condition directly into the RL reward, where the decrease condition encodes a desired minimal convergence rate, providing meaningful intermediate rewards. Although the idea may seem similar to potential shaping \cite{jeon2023benchmarking}, there are fundamental differences. Potential shaping replaces the original reward, while we propose an additional reward, and this addition is designed directly according to the structure of a valid CLF. Further, our proposed framework reduces all the tracking to a single Lyapunov function, which as we discuss later, experimentally, does not work well with potential shaping. Overall, the proposed reward formulation is principled, easy to integrate, and less sensitive to manual reward balancing than conventional tracking formulations.


Building on these properties, we propose a structured reward shaping approach that combines model-based reference planning with CLF-inspired objectives to guide policy learning toward stable and robust behaviors (Fig. \ref{fig:hero_fig}). We use two complementary planners for generating velocity-conditioned reference trajectories: (1) a reduced-order linear inverted pendulum (LIP) model for online generation, and (2) a precomputed hybrid zero dynamics (HZD) gait library from full-body offline trajectory optimization. Both produce periodic orbits but differ in fidelity and real-time suitability. From these trajectories, we construct a CLF \( V = \eta^\top P \eta \), where \( \eta \) denotes the output tracking error. Once a reference trajectory is available, implementing the CLF decrease condition requires minimal effort: the tuning process is straightforward, involving far fewer reward terms than conventional training pipelines and  resembling gain tuning in a tracking controller. The CLF and its associated stability condition is embedded into the RL reward to shape the policy to promote stable behaviors during training. We validate our approach through both simulation and hardware experiments on the Unitree G1 humanoid robot. Results demonstrate that CLF-based reward shaping improves tracking performance, reduces variance under randomized model perturbations, and enhances robustness  compared to standard RL baselines—offering a promising path toward more robust and theoretically grounded locomotion learning.

\begin{figure*}
    \centering
    \includegraphics[width=\linewidth]{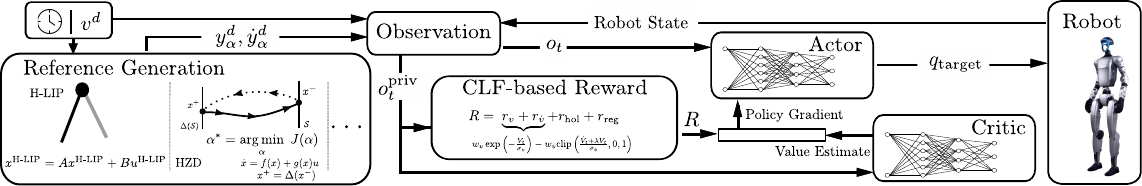}
    \caption{Overview of CLF-guided reinforcement learning framework. A desired velocity \( v^d \) is passed to a reference generator (e.g., H-LIP or HZD) to produce targets \( y^d_\alpha, \dot{y}^d_\alpha \). These, along with the robot state and privileged variables \( o_t^\text{priv} \), are used to compute a CLF-based reward. The actor-critic policy is trained with this reward and outputs joint targets \( q_\text{target} \) for the robot.}
    \label{fig:overview}
    \vspace{-5mm}
\end{figure*}

%% file: sections/background.tex
\section{Preliminaries}
Consider a robotic system with generalized coordinates \( q = [q_b^\top,\, q_a^\top]^\top \in \mathcal{Q} \subset \mathbb{R}^n \), where \( q_b \in SE(3) \) denotes the floating-base coordinates and \( q_a \in \mathbb{R}^m \) the actuated degrees of freedom (DoFs). The control input is \( u \in \mathbb{R}^m \), and the full-order state is given by \( x = [q^\top,\, \dot{q}^\top]^\top \).
Legged locomotion is often modeled as a hybrid system with continuous dynamics and discrete transitions such as impact events across domains \cite{westervelt2003hybrid}. 
Let $\mathcal{D}$ denote the domain and $\mathcal{S}$ the guard of the system. The hybrid system $\mathcal{H}$ can be described as:
\begin{numcases}{\mathcal{H}:}
\dot{x} = f(x) + g(x)\,u & $x \in \mathcal{D} \setminus \mathcal{S}$, \label{eq: continuous_dynamics}
\\
x^+ = \Delta(x^-) & $x^- \in \mathcal{S}$, \label{eq: discretecontrol}
\end{numcases}
where \eqref{eq: continuous_dynamics} represents the continuous full-order Lagrangian dynamics, while $\Delta: \mathcal{S} \rightarrow \mathcal{D}$ represents the discrete event when contact mode changes, and the superscripts ``$-$'' and ``$+$'' indicates the moments immediately before and after the discrete impact event, respectively. 

Within each domain, the system dynamics can be derived from the Euler-Lagrange equation:
\begin{align}
  D(q)\ddot{q} + H(q,\dot{q}) = B u + J_h(q)^TF \label{eq:dynamics} \\
  J_h(q) \ddot{q} + \dot{J}_h(q,\dot{q})\dot{q} = 0 \label{eq:hol_dynamics}
\end{align}
where $D(q):\mathcal{Q}\to \mathbb{R}^{n\times n}$ is the mass-inertia matrix, $H: \mathcal{Q} \times T\mathcal{Q} \to \mathbb{R}^n$ contains the Coriolis and gravity terms, and $B\in\mathbb{R}^{n\times m}$ is the actuation matrix. The Jacobian of the holonomic contact constraint is $J_h(q) \in \mathbb{R}^{h \times n}$ and the associated constraint wrench is $F \in \mathbb{R}^h$.

\subsection{Periodic Gait Generation}
Legged locomotion is fundamentally periodic, alternating between stance and swing phases. This natually motivates a step-to-step (S2S) analysis via the discrete-time Poincaré return map $P: \mathcal{S} \to \mathcal{S}$, with $x_{k+1} = P(x_k)$ giving the state from one impact to the next. A periodic gait is a fixed point $x^*$ such that $P(x^*) = x^*$ and is exponentially stable when this fixed point is locally attracting \cite{morris2005restricted}. Periodic motions can thus be synthesized by shaping the return map to yield a stable fixed point. Here, we focus on two complementary approaches: hybrid zero dynamics (HZD) and H-LIP models.

\noindent \textit{\underline{Hybrid Zero Dynamics:}} 
The Hybrid Zero Dynamics (HZD) approach synthesizes stable bipedal gaits by enforcing a set of phase-based virtual constraints, typically specified using Bézier polynomials with coefficients $\alpha$. These constraints define desired trajectories for joint or end-effector motion as functions of a monotonically increasing phase variable. Driving the output of the virtual constraints to zero renders the zero dynamics manifold, denoted $\mathcal{Z}^\alpha$, invariant and attractive. To achieve periodic walking under hybrid dynamics, the manifold must also be impact invariant, meaning that the system returns to $\mathcal{Z}^\alpha$ after a discrete event (e.g., foot impact):
\begin{align}\label{eq:impact_invariance}
    \Delta(\mathcal{Z}^{\alpha} \cap \mathcal{S}) \subset \mathcal{Z}^{\alpha}
\end{align}
where $\Delta$ denotes the reset map and $\mathcal{S}$ is the switching surface. When this condition holds, the system implicitly stabilizes a fixed point in the step-to-step dynamics.

\noindent \textit{\underline{Hybrid Linear Inverted Pendulum:}}
Assuming constant (centroidal) angular momentum and fixed \(\compos^{z}\) yields the well-known Linear Inverted Pendulum (LIP) for humanoid walking \cite{kajita2003biped,kajita2002realtime}. To capture the hybrid nature of the gait, \cite{xiong20223} introduced the Hybrid LIP (H-LIP), which, like LIP, assumes constant CoM height and step timing, but explicitly models single  (SSP) and double-support (DSP) domains with a reset map. The mismatch to the full-order dynamics is treated as a bounded disturbance, enabling state-feedback foot-placement laws that render the horizontal CoM states bounded and invariant. 
Assuming no velocity jump, the two domains with step size $u^{\textrm{H-LIP}}$ compose into the equivalent reset:\begin{numcases}{\Delta_{\textrm{SSP}^- \rightarrow \textrm{SSP}^+}:} 
\comvel^+ = \comvel^- \nonumber\\ \compos^+ = \compos^- + \comvel^-T_{\textrm{DSP}}- u^{\textrm{H-LIP}}.\nonumber \end{numcases} The S2S dynamics then take the linear form $x_{k+1}^{\textrm{H-LIP}} = Ax_{k}^{\textrm{H-LIP}} + Bu^{\textrm{H-LIP}}_{k}$, with $x^{\textrm{H-LIP}} = [\compos,\comvel]^\top$, which serves as a reduced-order approximation of the Poincaré map of the full-order system; see \cite{xiong20223} for derivation details.

\noindent \textit{\underline{Remarks:}}
Although H-LIP enables online, feedback-based foot placement, we use steady-state trajectories derived by fixed-point identification and analytic rollout. Regardless of model order (full-order HZD vs. reduced-order H-LIP), both methods stabilize periodic gaits by regulating the Poincaré map, motivating our reward-shaping framework.

\subsection{Control Lyapunov Functions (CLFs)}
For a nonlinear control-affine system of the form $\dot{x} = f(x) + g(x)u$, a continuously differentiable, positive definite function $V(x) : \mathbb{R}^{2n} \rightarrow \mathbb{R}$ is called an exponentially stabilizing control Lyapunov function (CLF) if there exists a control input $u$ such that, for some scalar $\lambda > 0$,
\begin{align}
    \inf_{u}\nabla_x V(x)(f(x) + g(x)u)  < -\lambda V(x).
    \label{eq:clf_condition}
\end{align}
This inequality guarantees that $V(x)$ decreases along system trajectories, implying exponential convergence of the state $x$ to the origin (or a desired manifold). Importantly, the CLF condition \eqref{eq:clf_condition} defines a set of admissible control inputs rather than a unique control law. Any control input $u$ that satisfies the inequality contributes to stability. This property makes CLFs especially appealing for reinforcement learning: rather than enforcing a fixed feedback controller, we can embed the CLF condition directly into the reward function, encouraging the policy to learn Lyapunov-stable behavior, without constraining it to a specific control action.
For hybrid systems such as bipedal locomotion with discrete impact events, CLFs can be applied only within each continuous domain of the dynamics. Prior work has shown that if the rate of convergence in the continuous phase is sufficiently fast, it can offset the destabilizing effects of impacts \cite{ames_rapidly_2014}, providing certifiable stability.

%% file: sections/reward_design.tex
\section{Reference-Guided Reward Shaping}
We consider the reinforcement learning (RL) problem of learning a locomotion policy \(\pi_\theta(a_t \vert o_t)\) that maps proprioceptive observations \(o_t\) to actions \(a_t\) using Proximal Policy Optimization (PPO) \cite{schulman2017proximal}. Instead of relying solely on handcrafted rewards, we embed model-based prior knowledge into the reward through a control Lyapunov function (CLF), promoting stability with respect to a reference trajectory. The full framework, including reference generation and CLF-based reward shaping, is illustrated in Fig. \ref{fig:overview}.

\subsection{Reference Trajectory Generation}
Our approach assumes a nominal reference trajectory that encodes a periodic or quasi-periodic walking motion given a desired velocity. We focus on two sources of reference trajectories: the H-LIP model and full-order HZD optimization, but the framework is agnostic to the origin of the trajectory.

\noindent \textit{\underline{H-LIP-Based Reference:}}  
Given a target walking velocity \( v^d \), we first solve for the H-LIP fixed point \( x^* = A x^* + B u^* \), where the step size \( u^* = v^d T \) corresponds to the desired walking speed and nominal step time. This yields a periodic center-of-mass (CoM) state \( x^* = [p_{\text{com}}, v_{\text{com}}]^\top \) consistent with the H-LIP S2S dynamics. Using this fixed point, we analytically generate the CoM trajectory over one step. The swing foot trajectory is constructed using a time-parameterized 5th-order Bézier curve to ensure foot clearance and smooth touchdown. For other end-effector or joint trajectories, we use simple sinusoidal patterns that are synchronized with the gait phase to produce time-varying desired trajectories.

\noindent \textit{\underline{HZD-Based Reference:}}  
To obtain HZD trajectories, we solve an offline trajectory optimization problem to determine the Bézier coefficients $\alpha$ that define the virtual constraints:
\begin{align*} 
   \{\alpha^*,X^*\} = & \argmin_{\alpha,X}    \Phi(X) \\
    \text{s.t.}\quad 
    & \dot{x} = f(x) + g(x)u \tag{Dynamics} \\
    & \Delta(\mathcal{S} \cap \mathcal{Z}_\alpha) \subset \mathcal{Z}_\alpha \tag{HZD Condition} \\
    & X_{\text{min}}  \leq X \leq X_{\text{max}} \tag{Decision Variables} \\
    & c_{\text{min}}  \leq c(X) \leq c_{\text{max}} \tag{Physical Constraints},
\end{align*}
where $X = (x_0,u_0...,x_N,u_N) \in \R^{n_d}$ denotes the collection of decision variables, with $n_d \in \mathbb{N}$. Each $x_i$ represents the state at node $i$. The cost function is given $\Phi: \R^{n_d} \to \R_{\geq 0}$, while the constraints are represented by $c: \R^{n_d} \to \R^{n_p}$ with $n_p \in \mathbb{N}$. These constraints encode the physical laws governing locomotion, such as friction cone conditions, workspace limits, and actuator capacity \cite{reher2020algorithmic}. Solving this optimization yields a stable periodic solution to the walking dynamics for a single step, parameterized by a fixed set of B\'ezier coefficients $\alpha^*$. This solution is symmetrically remapped to produce a full gait cycle with alternating stance legs, yielding reference trajectories for the complete walking motion.

\subsection{CLF-Based Reward Design}
Let $\eta(t)$ denote the output tracking error between the current state and the reference trajectory:
\begin{equation}
\eta(t) =
\begin{pmatrix}
y^{d}(t) - y(q)\\
\dot y^{d}(t) - \dot y(q,\dot q)
\end{pmatrix},    
\end{equation}
and the corresponding control Lyapunov function is
\begin{align}
V(\eta) = \eta^\top P \eta.
\end{align}
The matrix \( P \succ 0 \) is the unique solution to the Continuous-Time Algebraic Riccati Equation (CARE) for the feedback linearized model (see below). Prop. \ref{prop:clf} proves that this is a valid CLF for the nonlinear system in single support flat-footed walking.

\begin{proposition} [CLF Existence]
    \label{prop:clf}
Consider the fully-actuated humanoid system given by Eq. \eqref{eq:dynamics} and satisfying the constraint given by Eq. \eqref{eq:hol_dynamics} and friction cone constraints. If $Q \succ 0$ in the CARE, and the jacobian of the virtual constraints $\frac{\partial y}{\partial q}$ are invertible, then the function $V(\eta) = \eta^T P \eta$ is a valid CLF for the full nonlinear dynamics during the single support phase.
\end{proposition}
\begin{proof}
    If the holonomic constraints, Eq. \eqref{eq:hol_dynamics}, are satisfied, and the friction cone is respected, then we can write the system dynamics without the constraint Jacobian \cite{grizzle_3d_2010}:
    $
        D(q)\ddot{q} + H(q, \dot{q}) = Bu.
    $
    Define the virtual constraints error as $\bar{y} = y^d(t) - y(q)$ which we want to drive to $0$. Differentiating twice yields
    \begin{align*}
        \ddot{\bar{y}} = \underbrace{\frac{\partial L_f \bar{y}}{\partial q}\dot{q} + \frac{\partial \bar{y}}{\partial q} \left( D(q)^{-1} (-H(q, \dot{q}) \right)}_{L_f^2 \bar{y}(q, \dot{q})}  + \underbrace{\frac{\partial \bar{y}}{\partial q} \left( D(q)^{-1} Bu \right)}_{L_gL_f \bar{y}(q, \dot{q})u}.
    \end{align*}
    From the fully actuated assumption, $\text{dim}\: q = \text{dim}\: u = \text{rank}\: B$ \cite{grizzle_3d_2010}. Therefore $L_g L_f \bar{y}(q, \dot{q})$ is invertible under the assumptions of the proposition. We can now apply the feedback linearizing controller: $u = K(x)$, where $K(x)$ is defined as
   $
        K(x) := L_gL_f \bar{y}(q, \dot{q})^{-1} \left(-L_f^2 \bar{y}(q, \dot{q}) + v \right)
$
    where $v$ is a new, auxiliary, input. This leads to the closed loop system $\dot{\eta} = A_{\eta} \eta + B_{\eta} v$ with $A_\eta$ and $B_\eta$ defined as
  $
        A_\eta = \begin{bmatrix} 0 & I \\ 0 & 0 \end{bmatrix}\quad B_\eta = \begin{bmatrix} 0 \\ I \end{bmatrix}.
$
    This is a double integrator system. We solve the CARE for this system and use the resulting $P$ to synthesize an LQR gain: $K_\eta = R_\eta^{-1} B_\eta^T P$. We can then define $v = K_\eta \eta$ as an optimal stabilizing auxiliary controller. Denote the closed loop $\eta$ system as $A_\eta^{\text{cl}} = A_\eta + B_\eta K_\eta$. Consider the function $V(\eta) = \eta^T P \eta$. Using the feedback controller and differentiating yields
    \begin{equation}
        \dot{V}(\eta) = \frac{\partial V}{\partial \eta} \dot{\eta} = 2\eta^T P A_\eta^{\text{cl}}\eta = \eta^T (A_\eta^{\text{cl}T} P + P A_\eta^{\text{cl}})\eta.
    \end{equation}
    Then, using the CARE, we get that $A_\eta^{\text{cl}T} P + P A_\eta^{\text{cl}} = -Q - PB_{\eta}R_{\eta}^{-1}B_{\eta}^TP := \bar{Q}$. By assumption $Q$ is positive definite and note that the other term is positive semi-definite. Therefore
    \begin{equation}
        \dot{V}(\eta) \leq -\lambda_{\text{min}}(\bar{Q})\|\eta\|^2 \leq -\frac{\lambda_{\min}(\bar{Q})}{\lambda_{\min}(P)}V(\eta)
    \end{equation}
    where $\lambda_{\min}$ denotes the minimum eigenvalue. Using $V(\eta)$ as a Lyapunov function, the set $\{ u \: | \: \dot{V} \leq -\lambda V \}$ is non-empty and thus $V(\eta)$ is a CLF for the nonlinear system.
\end{proof}

Let $\sigma_{v} = \mu_{\max}(P)\,\eta_{\max}^2$, where $\mu_{\max}(P)$ is the maximum eigenvalue of the CLF matrix $P$ and $\eta_{\max}$ is an empirical bound on the tracking error. With $V_t = V\bigl(\eta(t)\bigr)$, the CLF tracking reward is then
\begin{align}
r_{v} &= w_{v} \exp\!\left(-\frac{V_t}{\sigma_{v}}\right).
\tag{CLF Tracking}\label{eq:clf_tracking}
\end{align}
We formulate the CLF decay reward using a clip function:
\begin{align}
r_{\dot{v}}^{\text{clip}} &= -w_{\dot{v}} \;\operatorname{clip}\!\left(\tfrac{\dot{V}_t + \lambda V_t}{\sigma_{\dot{v}}},\,0,\,1\right) 
\tag{CLF Decay}\label{eq:clf_decrease_clip},
\end{align}
where $\operatorname{clip}(x,a,b) := \min(\max(x,a),b)$ and \( \sigma_{\dot{v}} = 2 ||P|| \eta_{\max}\dot{\eta}_{\max} + \lambda \mu_{\max}\eta_{\max}^2\) and \( \lambda > 0 \) specifying the desired CLF decay rate. Specifically, this reward is providing a penalty if the CLF is not decaying sufficiently, otherwise there is no reward. Note that during RL training, to avoid explicitly computing \( \dot{\eta} \) and \( \dot{V} \), we approximate the Lyapunov derivative using finite differences:
$
\dot{V}_t \approx \frac{V_{t+1} - V_t}{\Delta t}
$.


\subsection{Auxiliary Rewards and Final Reward Composition}
In addition to the CLF terms, we incorporate several auxiliary rewards commonly used in RL for legged systems. To enforce stance-foot holonomic constraints, we define a combined reward:
\begin{equation*}
r_{\mathrm{hol}} = 
w_{\mathrm{hpos}} \exp\left(-\frac{\|p_{\mathrm{st}} - p_{\mathrm{st}}^0\|}{\sigma_p}\right) 
+ w_{\mathrm{hvel}} \exp\left(-\frac{\|v_{\mathrm{st}}\|}{\sigma_v}\right),
\end{equation*}
where \(p_{\mathrm{st}}\) and \(p_{\mathrm{st}}^0\) are the current and the initial stance foot positions when it enters the domain, \(v_{\mathrm{st}}\) is the stance foot velocity, and \( w_{\mathrm{hpos}}\) and \(w_{\mathrm{hvel}} \) are weights.

 To discourage excessive control effort, abrupt actions, and joint limit violations, we include a regularization reward:
\begin{align*}
r_{\mathrm{reg}} =
- &w_u \|u_t\|^2 
- w_{\Delta a} \|a_t - a_{t-1}\|^2 \notag \\
 - &w_{q_{\text{limit}}}\left\| 
\max\left(0, q_{\min} - q_t\right) + 
\max\left(0, q_t - q_{\max}\right) 
\right\|_1,
\end{align*}
where each term is weighted by its corresponding coefficient to balance control effort, motion smoothness, and adherence to joint limits.

The final reward used for policy training is the weighted sum of all components:
$
R =\;
r_{v} 
+ r_{\dot{v}} 
+ r_{\mathrm{hol}}
+ r_{\mathrm{reg}}.
$
This reward structure encodes both task performance and physical feasibility, allowing the policy to balance tracking, stability, and regularization objectives. By combining model-based references (e.g., from reduced-order planning or offline trajectory optimization) with CLF-inspired reward shaping and constraint-aware regularization, our framework embeds control-theoretic structure into the reinforcement learning process. This guides training toward stable, robust, and physically plausible locomotion, while preserving the flexibility and adaptability of model-free RL.

%% file: sections/framework.tex
\subsection{Implementation Details}
We demonstrate our framework on the 29-DoF Unitree G1 humanoid robot, planning over 21 actuated DoF. The hand and wrist joints are fixed during training and controlled with zero desired position and velocity during deployment.

\noindent \textit{\underline{Reference Generation:}} 
We use the same set of end-effector and joint positions for both HZD and H-LIP-based trajectories. The full reference output \( y_{\alpha}^d(t) \in \mathbb{R}^{n_y} \), with \( n_y = 21 \) is structured as follows:
$
y_{\alpha}^d =
\begin{pmatrix}
p^d_{\mathrm{com}} &
\phi^d_{\mathrm{pelvis}} &
p^d_{\mathrm{sw}} &
\theta^d_{\mathrm{sw}} &
q^d_{\mathrm{shoulder}} &
q^d_{\mathrm{elbow}}
\end{pmatrix}^\top$
where the terms represent the desired CoM position, pelvis orientation, swing foot pose, and arm joint angles. One can verify that this set of outputs lead to an invertible $\frac{\partial \bar{y}}{\partial q}$ that satisfy Prop \ref{prop:clf}. To account for turning during training, we heuristically adjust the reference trajectory based on the commanded angular velocity. Specifically, we modify the yaw orientation of all end-effector frames to align with the integrated yaw derived from the commanded angular velocity over time. The HZD optimization is computed offline using IPOPT \cite{wachter_implementation_2006} and Casadi \cite{andersson_casadi_2019}. Hard constraints are used to enforce dynamics, periodicity, virtual constraints, and step length among others. The optimization problem is formulated as a multiple-shooting problem over a single swing phase. 

\noindent \textit{\underline{Policy Structure:}}  
The policy receives a combination of proprioceptive and task-relevant inputs, including angular velocity, projected gravity, commanded linear and angular velocities, relative joint positions and velocities, the previous action, and a phase-based time encoding using \( \sin\left(2\pi t / t_{\mathrm{period}}\right) \) and \( \cos\left(2\pi t / t_{\mathrm{period}}\right) \). The stepping period $t_{\mathrm{period}}$ is set to \(0.8\,\mathrm{s}\), corresponding to a full gait cycle (i.e., two steps). The policy outputs joint position commands relative to a default symmetric standing pose at 50\,Hz. This default pose is fixed across all policies we trained. Both the actor and critic networks use a fully connected feedforward architecture with hidden layer dimensions of [512, 256, 128] and
Exponential Linear Unit activation function. The critic receives additional privileged information $o^{\mathrm{priv}}_t$, such as stance and swing foot linear and angular velocities, reference trajectory positions and velocities, and binary contact state indicators. 

\begin{table}[t]
\centering
\caption{Reward weights used for HZD-CLF and LIP-CLF.}
\label{table:reward_coef}
\begin{tabular}{ll}
\toprule
\textbf{Reward Term}              & \textbf{Weight} \\
\midrule
Torque penalty \( w_\tau \)       & $1\times10^{-5}$ \\
Action-rate penalty \( w_{\Delta a} \) & $1\times10^{-3}$ \\
Joint limit penalty \(w_{q_{\mathrm{limit}}}\) & $1.0$\\
CLF tracking reward \( w_v \)     & $10.0$ \\
CLF decay penalty \( w_{\dot{v}} \) & $2.0$ \\
Holonomic position reward \( w_{\text{hpos}} \) & $4.0$ \\
Holonomic velocity reward \( w_{\text{hvel}} \) & $2.0$ \\
\bottomrule
\end{tabular}
\vspace{-5mm}
\end{table}

\noindent \textit{\underline{Training Procedure:}}
We use \textit{IsaacLab}, a GPU-accelerated simulation framework built on top of NVIDIA Isaac Sim, with its development initiated from the \textit{Orbit} framework~\cite{mittal2023orbit}. For policy training, we use the PPO implementation from Robotics System Lab RL library~\cite{rudin2022learning}.

All training, including baseline and CLF-based results, use the same set of terrain settings, robot models, and domain randomization parameters. To improve policy robustness and facilitate transfer to hardware, we apply domain randomization to physical properties such as link masses, static and dynamic friction coefficients, and the center of mass position. Additionally, external perturbations are introduced by applying randomized velocity impulses in the \(x\) and \(y\) directions to the base link at fixed time intervals.
Training is conducted over a range of commanded velocities, with linear velocity \( v_x \in [-0.75, 0.75] \) and yaw rate \( \omega_z \in [-0.5, 0.5] \). Given a commanded velocity, $v^d$, the reference trajectory is either retrieved from the precomputed gait library by selecting the closest matching forward velocity in the case of HZD, or generated online using the analytic form of the H-LIP gait library, which provides closed-form solutions across the velocity space. For both versions, we used the same reward listed under Table \ref{table:reward_coef}.

The custom baseline policy is trained using a manually designed reward function composed of multiple heuristic terms. These include tracking linear velocities in the $x$ and $y$ directions as well as angular yaw velocity, while penalizing vertical velocity, angular velocities about the $x$ and $y$ axes, joint accelerations, joint velocities, joint torques, and abrupt changes in actions. The reward also discourages deviations from an upright base orientation and enforces compliance with joint limits. Additional terms promote behaviors such as maintaining an upright posture, preventing foot slip during contact, preserving nominal hip and torso positions, ensuring sufficient foot clearance, and regulating contact timing. Finally, the policy is penalized for deviations of the arm and torso joints from predefined reference configurations.

%% file: sections/results.tex
\section{Results}
\begin{figure}
    \centering
    \includegraphics[width=1.0\linewidth]{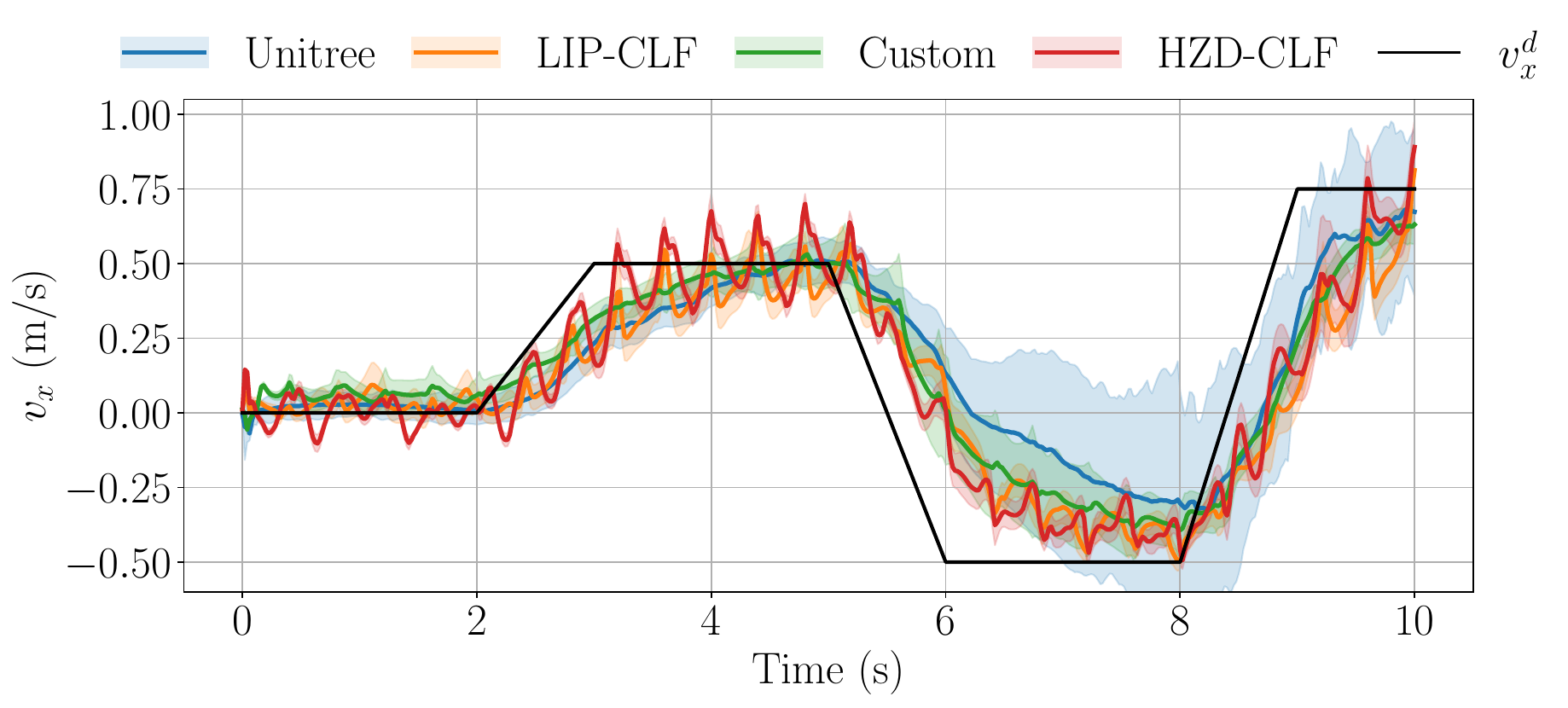}
    \caption{Tracking performance with torso mass randomly displaced within a box of size $\pm [0.05\text{ (x)}, 0.05\text{ (y)}, 0.01\text{ (z)}]$m around the nominal location. Fifty displacements are uniformly sampled, and the resulting mean and standard deviation of performance for each policy are plotted.}
    \label{fig:mass_randomization}
\end{figure}

\begin{table}[h]
\centering
\caption{Tracking Velocity Performance Comparison.}
\label{table:clf_comparison}
\begin{tabular}{llcc}
\hline
\textbf{Coord.} & \textbf{Method} & \textbf{Vel. Err. Mean} $\downarrow$ & \textbf{Vel. Err. Std} $\downarrow$ \\
\hline
\multicolumn{4}{c}{\textbf{Center of Mass (cm/s)}} \\
\hline
\multirow{2}{*}{$x$} 
& CLF (ours) & \textbf{7.3} \textit{(-24.4\%)} & \textbf{4.6} \textit{(-20.1\%)} \\
& Tracking only & 9.6 & 5.8 \\
\multirow{2}{*}{$y$}
& CLF (ours) & \textbf{5.9} \textit{(-4.6\%)} & \textbf{3.7} \textit{(-7.0\%)} \\
& Tracking only & 6.2 & 4.0 \\
\multirow{2}{*}{$z$}
& CLF (ours) & \textbf{7.2} \textit{(-7.3\%)} & \textbf{4.4} \textit{(-4.7\%)} \\
& Tracking only & 7.7 & 4.7 \\
\hline
\multicolumn{4}{c}{\textbf{Pelvis Orientation (rad/s)}} \\
\hline
\multirow{2}{*}{Roll}
& CLF (ours) & \textbf{0.141} \textit{(-4.7\%)} & \textbf{0.084} \textit{(-9.4\%)} \\
& Tracking only & 0.148 & 0.093 \\
\multirow{2}{*}{Pitch}
& CLF (ours) & \textbf{0.312} \textit{(-6.8\%)} & \textbf{0.125} \textit{(-2.6\%)} \\
& Tracking only & 0.335 & 0.128 \\
\multirow{2}{*}{Yaw}
& CLF (ours) & \textbf{0.319} \textit{(-8.5\%)} & \textbf{0.165} \textit{(-4.8\%)} \\
& Tracking only & 0.348 & 0.173 \\
\hline
\multicolumn{4}{c}{\textbf{Swing Ankle (cm/s)}} \\
\hline
\multirow{2}{*}{$x$}
& CLF (ours) & \textbf{17.2} \textit{(-18.4\%)} & \textbf{11.3} \textit{(-18.6\%)} \\
& Tracking only & 21.1 & 13.9 \\
\multirow{2}{*}{$y$}
& CLF (ours) & \textbf{9.7} \textit{(-13.3\%)} & \textbf{7.4} \textit{(-10.4\%)} \\
& Tracking only & 11.2 & 8.3 \\
\multirow{2}{*}{$z$}
& CLF (ours) & 21.2 \textit{(+3.3\%)} & \textbf{7.4} \textit{(-0.9\%)} \\
& Tracking only & \textbf{20.5} & 7.4 \\
\hline
\multicolumn{4}{c}{\textbf{Swing Ankle Orientation (rad/s)}} \\
\hline
\multirow{2}{*}{Roll}
& CLF (ours) & \textbf{0.346} \textit{(-7.6\%)} & 0.272 \textit{(+13.2\%)} \\
& Tracking only & 0.374 & \textbf{0.240} \\
\multirow{2}{*}{Pitch}
& CLF (ours) & 0.641 \textit{(+8.9\%)} & 0.445 \textit{(+4.6\%)} \\
& Tracking only & \textbf{0.589} & \textbf{0.425} \\
\multirow{2}{*}{Yaw}
& CLF (ours) & \textbf{0.228} \textit{(-5.2\%)} & \textbf{0.176} \textit{(-7.5\%)} \\
& Tracking only & 0.241 & 0.190 \\
\hline
\end{tabular}
\vspace{-5mm}
\end{table}

\begin{figure*}
    \centering
    \includegraphics[width=1.0\linewidth]{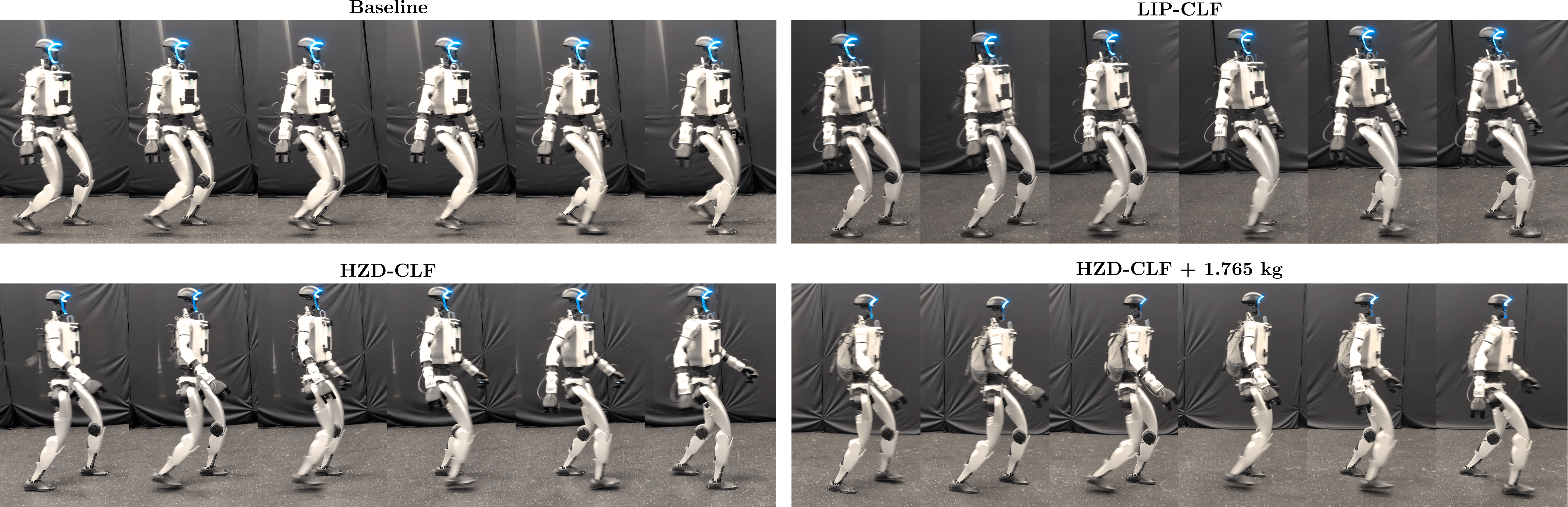}
    \caption{Snapshots of the three policies throughout a stride on the Unitree G1 robot. These images depict the walking motion in steady state walking with a commanded velocity of $v_x^d = 0.75$ m/s. }
    \vspace{-5mm}
    \label{fig:gait_tiles}
\end{figure*}

We evaluate four policy variants: a baseline with rewards from the Unitree open-source implementation\footnote{\url{https://tinyurl.com/e5nrjdb9}} (Unitree), a baseline trained with hand-tuned custom rewards (Custom), a H-LIP-based CLF shaped policy (LIP-CLF), and an HZD-based CLF-shaped policy (HZD-CLF)\footnote{Open source implementation is available at \url{https://github.com/Zolkin1/robot_rl}}. We also attempted to use potential shaping \cite{jeon2023benchmarking} to replace the tracking reward, but the policy never learned to walk. We hypothesize that potential shaping works better when some key rewards are left in their nominal form, but in our proposed framework there is only one tracking reward. The CLF-based policies were trained with many random seeds without any noticeable difference in performance, demonstrating that the proposed RL training pipeline is robust.
All policies are first validated in sim-to-sim transfer using MuJoCo \cite{todorov2012mujoco}. Because the Unitree-baseline underperforms compared to the hand-tuned baseline (Fig. \ref{fig:mass_randomization}), only three policies are tested on hardware (Fig. \ref{fig:gait_tiles}). The policies run onboard on an additional laptop mounted to the front of the torso during deployment, which also added additional mass (0.616 kg) to the robot. 

\noindent \textit{\underline{CLF Decay Condition:}}
To assess whether reference tracking alone is sufficient or if enforcing the CLF decay condition offers additional benefits, we conduct an ablation study. Table \ref{table:clf_comparison} compares two policies: one trained with only the tracking reward ($r_v$) and another trained with both the tracking and CLF decay rewards ($r_v$ and $r_{\dot{v}}$). We simulate 200 instances of the robot under out-of-distribution domain randomization to evaluate robustness. The mean and standard deviation of the tracking errors are reported at steady state while walking at 0.75 m/s. The results show that including the CLF condition generally reduces both the mean tracking error and its variability, averaging about a 10\% improvement in velocity tracking, and up to about 25\%.

\begin{figure}
    \centering
    \includegraphics[width=1.0\linewidth]{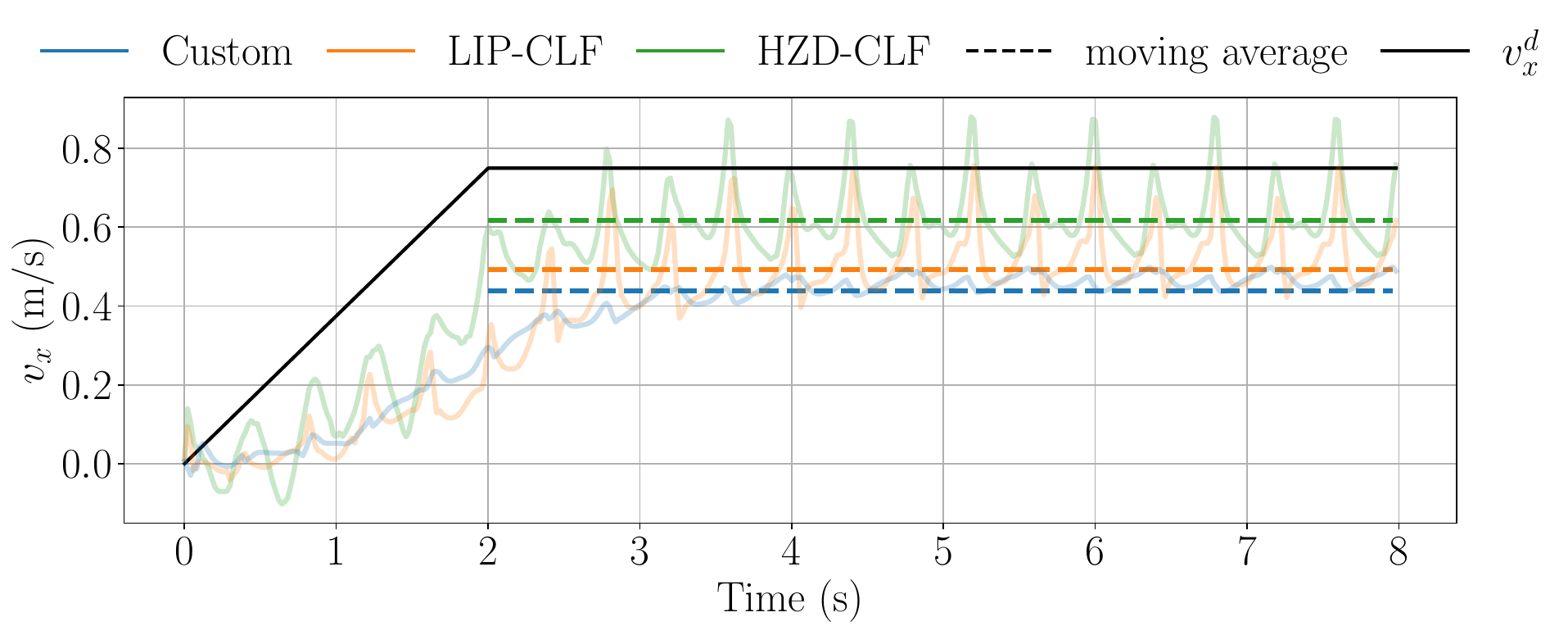}
    \caption{Robustness evaluation in simulation: HZD-CLF, LIP-CLF and the baseline policy are tested with an extra 8kg on the torso. A 2-second ramp commanded each controller up to the maximum training velocity. The steady-state mean velocities are indicated by dashed lines.}
    \label{fig:sim_added_mass}
    \vspace{-5mm}
\end{figure}

\noindent \textit{\underline{Robustness evaluation:}} 
To assess robustness to modeling mismatches, we introduce random perturbations to the torso mass location to simulate sim-to-real discrepancies in inertial properties. Specifically, the torso center of mass is uniformly displaced within a box of size $\pm [0.05~\text{m}~(x), 0.05~\text{m}~(y), 0.01~\text{m}~(z)]$. For each policy, we sample 50 randomized configurations and evaluate the velocity tracking performance across episodes, reporting the mean and standard deviation. As shown in Fig.~\ref{fig:mass_randomization}, both CLF-RL policies exhibit significantly lower performance variance compared to the hand-tuned baseline, with the unitree-baseline having the largest variance . This indicates improved robustness and consistency under structural uncertainties.
To further evaluate robustness, we introduce an 8\,kg payload to the torso and compare tracking performance across the three policy variants under a velocity ramp command. As shown in Fig.~\ref{fig:sim_added_mass}, the CLF-based RL policies maintain more accurate velocity tracking, demonstrating improved resilience to payload-induced dynamics changes.

\begin{figure}
    \centering
    \includegraphics[width=1.0\linewidth]{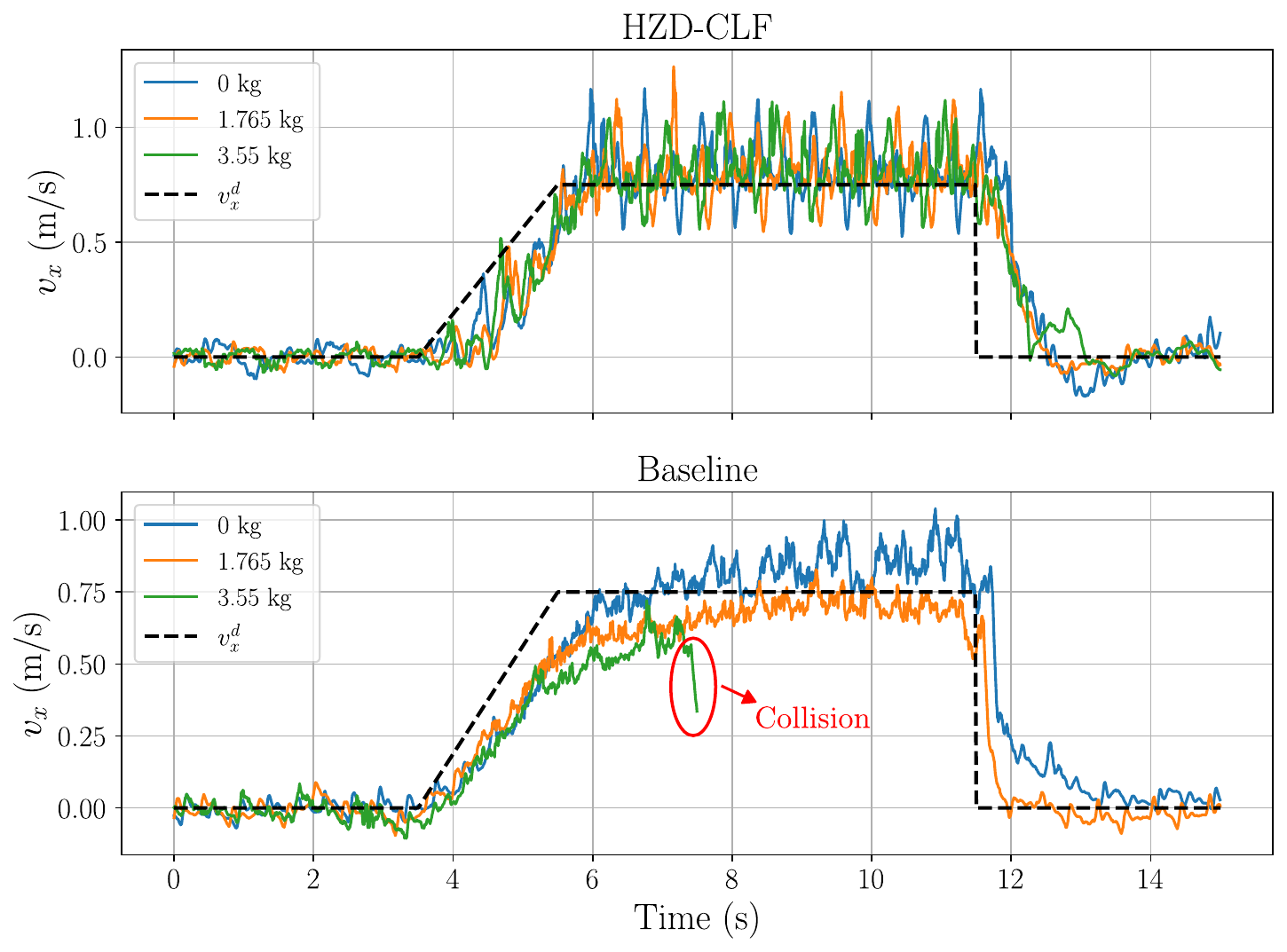}
    \caption{Hardware testing with added backpack mass. HZD-CLF maintains good tracking performance, while the baseline drifts off the walking course and collides with a table.}
    \label{fig:hardware_added_mass}
    \vspace{-5mm}
\end{figure}

\noindent \textit{\underline{Hardware Experiments:}} We deploy all the policies, except the Unitree baseline, on the Unitree G1 robot in a controlled indoor environment. Fig.~\ref{fig:gait_tiles} shows all three policies operating successfully on hardware. To quantify velocity tracking performance and robustness, we use a motion capture system to record global position and orientation data. This data is used solely for evaluation and is not fed into the controller. To remain consistent with simulation analysis, we track the pelvis frame and compute local-frame velocity by finite-differencing the position data (recorded at 240\,Hz) and aligning it with the robot’s yaw orientation. Fig.~\ref{fig:hardware_added_mass} demonstrates the improved robustness of the HZD-CLF policy relative to the baseline. To test this robustness, we attach a backpack to the robot and load it with either 1.765\,kg or 3.55\,kg of additional mass. The HZD-CLF policy maintains consistent tracking performance across all conditions, whereas the baseline policy exhibits significant degradation. In the heavier case, the baseline policy drifts off the walking course and fails to complete the test. These hardware results corroborate the simulation findings in Fig.~\ref{fig:mass_randomization}, which show that CLF-based policies exhibit lower performance variance under structural perturbations.

\begin{figure*}
    \centering
    \vspace{2mm}
    \includegraphics[width=1.0\linewidth]{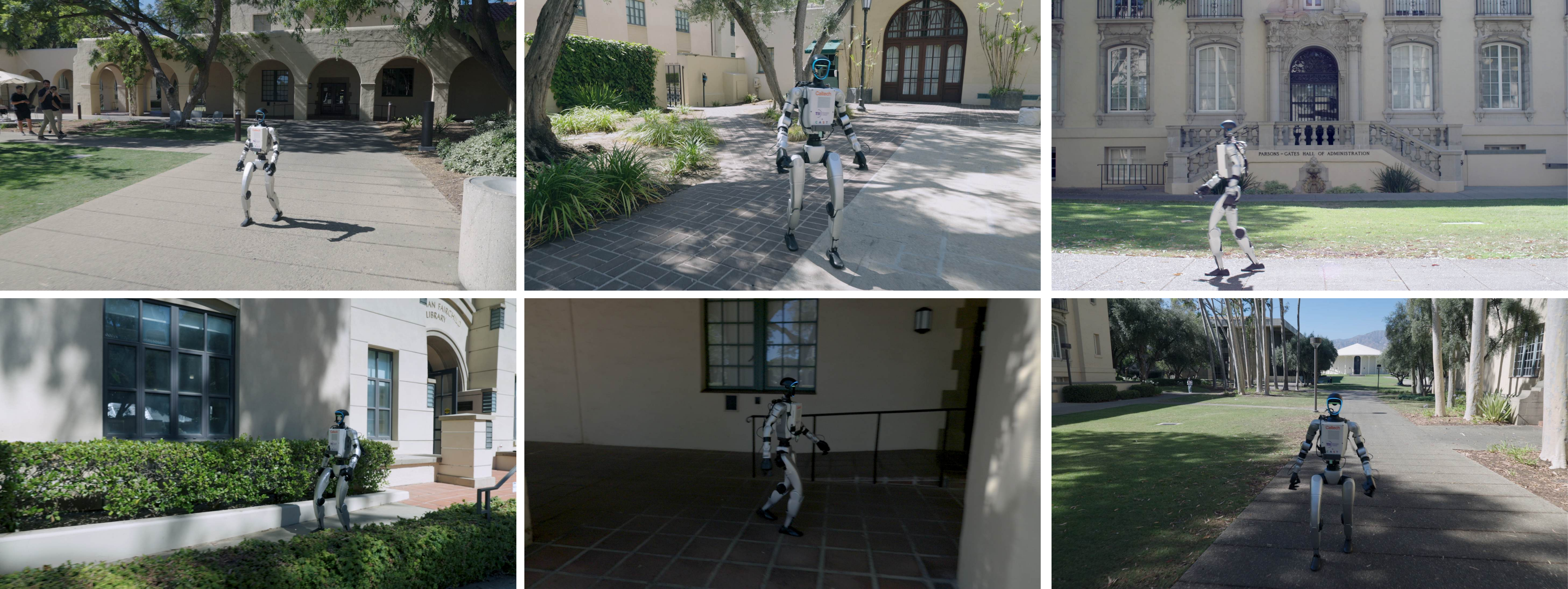}
    \caption{Demonstration of extensive outdoor testing of the HZD-CLF policy shows its ability to handle diverse flat-ground surfaces, including various tile and concrete types, as well as mild uphill and downhill slopes.}
    \label{fig:outdoor_experiments}
\end{figure*}

To evaluate the HZD-CLF policy under real-world conditions, we deploy the robot outdoors across a variety of environments. As shown in Fig.~\ref{fig:outdoor_experiments}, the test route includes diverse flat-ground surfaces such as concrete and tiled walkways, as well as mild slopes like ADA-compliant ramps. The robot traverses all terrain types within a single continuous 0.25-mile walking trial without any failures. The distance was determined by experimental design considerations, rather than reflecting any limitation of the policy.

